\documentclass[10pt,twocolumn,letterpaper]{article}

\usepackage{ICCV}              
\usepackage{booktabs}  
\usepackage{makecell}  
\usepackage{colortbl}
\usepackage{multirow}
\usepackage{tabularx}
\usepackage{subcaption}
\usepackage{amsmath}

\usepackage{amsfonts}
\usepackage{xcolor}         

\definecolor{colorFst}{HTML}{F59194}      
\definecolor{colorSnd}{HTML}{FAC791} 
\definecolor{colorThd}{HTML}{FFFF99}   
\definecolor{url_color}{RGB}{42, 83, 163}

\newcommand{\fs}[1]{\colorbox{colorFst}{\textbf{#1}}}
\newcommand{\nd}[1]{\colorbox{colorSnd}{\textbf{#1}}}     

\newcommand{\boldparagraph}[1]{\noindent{\bf #1} }

\definecolor{myRed}{rgb}{1.0, .0, .0}
\definecolor{myBlack}{rgb}{.0, .0, .0}
\definecolor{myBlue}{rgb}{0.0, .0, .8}

\usepackage{booktabs}
\newcommand{\method}{GaussianUpdate\xspace}

\definecolor{iccvblue}{rgb}{0.21,0.49,0.74}
\usepackage[pagebackref,breaklinks,colorlinks,allcolors=iccvblue]{hyperref}

\def\paperID{7565} 
\def\confName{ICCV}
\def\confYear{2025}

\title{GaussianUpdate: Continual 3D Gaussian Splatting Update \\ for Changing Environments}

\author{
Lin Zeng\footnotemark[1] \qquad
Boming Zhao\footnotemark[1]\qquad
Jiarui Hu \qquad
Xujie Shen \qquad
\\
Ziqiang Dang \qquad
Hujun Bao \qquad
Zhaopeng Cui\footnotemark[2]
\vspace{0.2cm}
\\
{State Key Lab of CAD \& CG, Zhejiang University}
\\
}
\begin{document}

\twocolumn[{
\renewcommand\twocolumn[1][]{#1}
\maketitle
\vspace{-7mm}
\includegraphics[width=1\linewidth, trim={0 3mm 0 0}, clip]{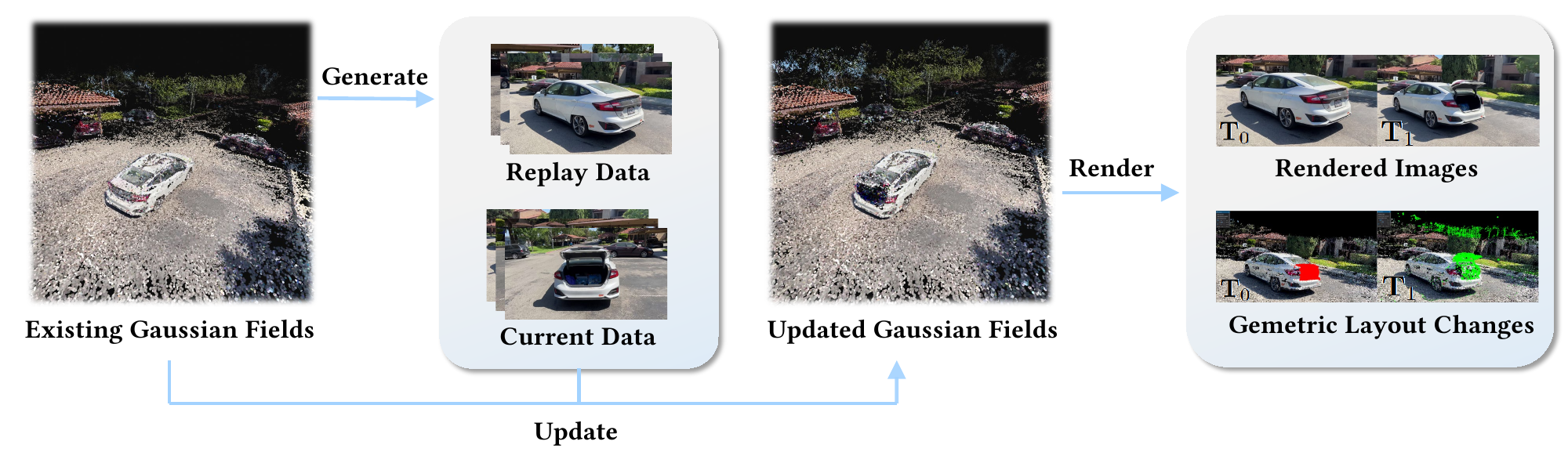}
\vspace{-7mm}
  \captionof{figure}{{\textbf{\method}. We present \method, a novel framework that can update the existing Gaussian fields by generatively replaying past data and currently captured data. The updated Gaussian fields can synthesize novel views across different times, capturing geometric layout changes between consecutive timestamps. 
  }
  \label{fig:pipeline}}
  \vspace{4mm}
}]

\maketitle
\renewcommand{\thefootnote}{\fnsymbol{footnote}}
\footnotetext[1]{\noindent Equal contribution.}
\footnotetext[2]{\noindent Corresponding author.}

\begin{abstract}
    Novel view synthesis with neural models has advanced rapidly in recent years, yet adapting these models to scene changes remains an open problem. Existing methods are either labor-intensive, requiring extensive model retraining, or fail to capture detailed types of changes over time. In this paper, we present \method, a novel approach that combines 3D Gaussian representation with continual learning to address these challenges. Our method effectively updates the Gaussian radiance fields with current data while preserving information from past scenes. Unlike existing methods, \method explicitly models different types of changes through a novel multi-stage update strategy. Additionally, we introduce a visibility-aware continual learning approach with generative replay, enabling self-aware updating without the need to store images. The experiments on the benchmark dataset demonstrate our method achieves superior and real-time rendering with the capability of visualizing changes over different times. Please refer to our project webpage for more informations: \urlstyle{tt}
      \textcolor{url_color}{\url{https://zju3dv.github.io/GaussianUpdate}}.
\end{abstract}
\\\\\\\\\    

\vspace{-30mm}
\section{Introduction}
\label{sec:Introduction}


High-quality scene reconstruction and realistic image rendering are crucial for many applications, such as AR/VR, 3D gaming, and robotics. 
In recent years, neural scene representations, represented by Neural Radiance Fields (NeRF)~\cite{nerf}, have demonstrated outstanding performance for novel view synthesis.
While these rendering results are impressive, the same scene often undergoes many changes over time. For example, the same room will have different objects, different furniture placements, and different lighting conditions at different times. Therefore, updating the original neural 
model is crucial to render images that align with the current appearance of the scene. 
Additionally, the knowledge from previous moments is valuable, and we hope to retain information from various times after updating the original 3D model. This would enable high-quality images of the same scene to be rendered at different moments when needed.

A simple approach is to retrain a neural model with the images from the current moment. However, preserving neural models from each moment would increasingly consume storage space. Another strategy is to directly update the neural model with data from the current time, but this may lead to catastrophic forgetting~\cite{catastrophic}, causing the neural model to lose appearance and geometric information from previous times and be unable to accurately render scenes from earlier times.
Existing methods~\cite{4d-gs, deformable-gs, 3dgstream, d-nerf, zhao2024gaussianprediction} for 4D modeling can represent continuous dynamics through dynamic fields, but they are not directly applicable to scenarios with abrupt scene changes, such as the sudden appearance or disappearance of objects.


Several attempts \cite{cl-nerf-iccv, cl-nerf-nips} have been made to integrate continual learning strategies \cite{continue-learning} into the learning of NeRF models.  For instance, CLNeRF~\cite{cl-nerf-iccv} combines continual learning with NeRF, updating the NeRF model with new data while preserving the geometric and appearance information from previous moments.  However, due to the inherent drawbacks, 
NeRF-based methods
fail to locate the changes over time and suffer from low rendering efficiency. 
Recently, 3D Gaussian Splatting ~\cite{3D-Gaussian} introduced a tile-based splatting rendering framework that uses 3D Gaussians to represent the scene. This explicit representation simplifies modeling and detection of fine-grained geometric and photometric changes, while enabling real-time rendering, offering a promising solution to the challenges mentioned earlier.

However, designing a system that combines 3D Gaussian representation with continual learning is far from straightforward and presents substantial challenges. Firstly, a temporal model is needed for the 3D Gaussian representation to address the differences in the appearance of the scene at different moments in time. Furthermore, when objects that originally existed in the scene are no longer present at the current moment, or new objects have appeared, we need a strategy to explicitly update the 3D Gaussian model. 
This process entails the precise removal of the 3D Gaussians associated with the objects that have disappeared and the addition of new 3D Gaussians at the necessary locations. The most relevant work is dynamic 3D Gaussian modeling~\cite{4d-gs, deformable-gs, gaufre}. These methods use an additional network to learn the deformation fields at different moments, warping the attributes of each 3D Gaussian from the canonical space to render temporally varying scene images. However, these methods assume that scene changes are caused by the continuous motion of objects and cannot model discrete changes in our tasks such as the appearance or disappearance of objects.

In this paper, to tackle these challenges, we introduce \method, a novel method that combines 3D Gaussian representation with continual learning for the first time. Our method effectively updates the previous neural model with current data while preserving as much information as possible from all past scenes. Compared to past moments, the changes in the current 3D scene are explicitly modeled as 
three types: lighting changes, disappearance of existing objects, and the addition of new objects. To model these discrete changes, we first propose a global appearance 
updating method for 3D Gaussians that utilizes hash encoding to store temporal information, thereby fitting the lighting variations at different moments. Additionally, we developed a 3D change detection method to identify and remove the 3D Gaussians corresponding to objects that have disappeared at the current moment. We also designed a simple but efficient COLMAP-based point addition strategy to represent newly added objects. 
With these meticulous designs, the proposed \method can ensure real-time photorealistic rendering of the scene's appearance at different times while also supporting the visualization of detailed changes over time.


Our contributions can be summarized as follows:
\begin{itemize}
    \item We present \method, a novel framework that innovatively integrates 3D Gaussian representation with continual learning, leveraging images of the current scene to update the previous model and render novel view images at different moments.
    \item We develop a novel multi-stage update strategy including hash encoding attribute updating, 3D change detection for object removal, and COLMAP-based point addition, which can adapt the previous 3D Gaussian model to the current changes in real-world scenes.
    \item We present the visibility-aware continual learning with generative replay which enables self-aware updating to various types of changes without the need to store additional images.
    \item Experimental results demonstrate our method achieves superior rendering quality compared to existing methods while ensuring real-time rendering and visualization of changes over different times.
\end{itemize}
\section{Related Work}
\label{sec:Related}

\boldparagraph{NeRF and 3D Gaussian Splatting.} 
Neural Radiance Field(NeRF) possesses robust multi-view synthesis capabilities. The vanilla NeRF~\cite{nerf} uses multi-layer perceptrons (MLPs) to map the input 3D coordinates and view direction into different colors and opacities and then employs volume rendering~\cite{volume-rendering} to produce images from novel views of the scene. However, NeRF can only reconstruct static scenes~\cite{hu2023cp}. To capture the dynamic changes in a scene, dynamic NeRF~\cite{d-nerf, K-planes, hexplane, TiNeuVox} trains a deformable field that deforms information from the canonical space to match different scene changes over time.
NeRF-w~\cite{nerf-w} and its follow up works~\cite{NeRF-on-the-go, factorized-NeRF} aiming to reconstruct from an internet collection dataset for outdoor benchmarks, introduces transient encoding and appearance encoding to capture variations in lighting and geometry at different times, thereby distilling the structural essence of the scene. However, the primary goal of NeRF-w's transient encoding is to filter out transient objects that occlude the main structure at different times. In contrast, our goal is to capture the changes at each moment, thereby rendering images for each specific instance.

Another major drawback of NeRF is its slow inference speed. Recently, 3D Gaussian Splatting~\cite{3D-Gaussian} employs a significant number of explicit 3D Gaussians to represent a static 3D scene~\cite{hu2024cg} and achieves real-time inference through tile-based splatting rendering. Many methods~\cite{4d-gs, deformable-gs, Gaussian-Flow, 3dgstream, zhao2024gaussianprediction} have improved upon the dynamic NeRF based on 3D Gaussian, thus achieving higher rendering quality and inference speed. However, to the best of our knowledge, there are currently no methods based on 3D Gaussians that can solve our task.

\boldparagraph{Continual Learning.}
Continual learning refers to models continuing to learn from new data after completing initial training, in order to adapt to changes in the environment. However, naive training on new data can lead to catastrophic forgetting of the knowledge previously learned from historical data. 
One strategy involves incorporating regularization terms that are unrelated to historical data~\cite{kirkpatrick2017overcoming, li2017learning}, yet this method frequently exhibits subpar performance in practical applications. Alternatively, some methodologies mitigate forgetting by replaying historical data, such as memory replay~\cite{chaudhry2019tiny, lopez2017gradient}, which stores historical data in a dedicated memory pool, and generative replay~\cite{shin2017continual}, which utilizes a generative model to replay historical data. Moreover, certain techniques prevent forgetting by selectively freezing portions of the network's model while introducing new network modules~\cite{mallya2018packnet, serra2018overcoming}.~\cite{cl-nerf-iccv} and~\cite{cl-nerf-nips} integrate continual learning into NeRF.~\cite{cl-nerf-iccv} achieves continual learning for NeRF in real-world scene changes by replaying historical data. In contrast,~\cite{cl-nerf-nips} attains higher training efficiency with a small number of images through a lightweight expert adaptor and a conflict-aware knowledge distillation strategy. We introduce continual learning to 3D Gaussian Splatting. Compared to NeRF-based continual learning methods, our approach achieves faster inference speeds and higher quality in novel view synthesis.
\section{Method}
\label{sec:Method}

\begin{figure*}[t]
  \centering
  \includegraphics[width=1\linewidth, trim={0 0 0 0}, clip]{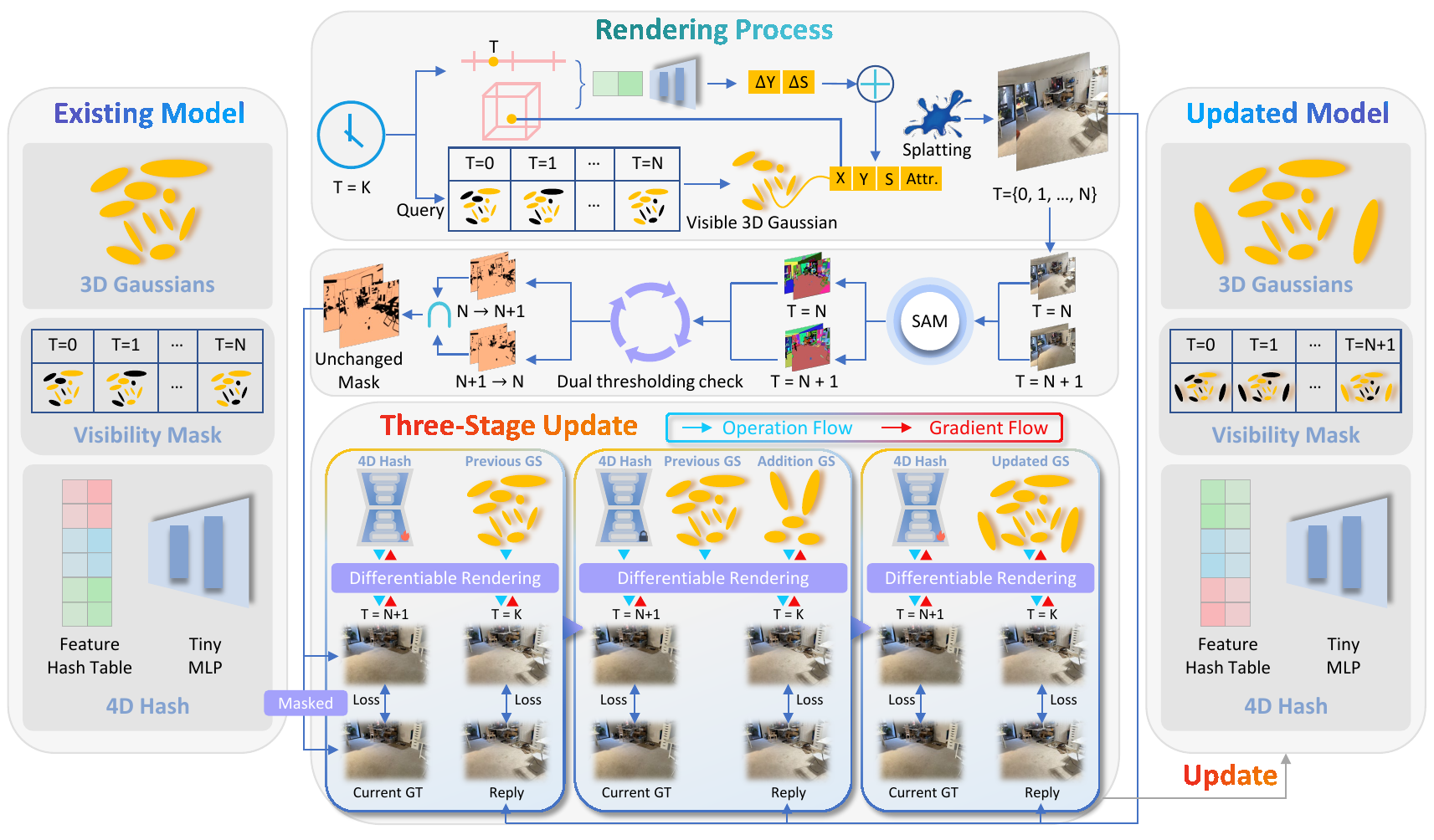}
  \caption{\textbf{System overview}. Given an existing 3D Gaussian Field, we use a visibility pool to record changes in scene layout and a 4D Hash as an appearance model to learn global illumination changes in the scene. The model update can be divided into three stages. In the first stage, we learn the global illumination changes in the layout-invariant regions. In the second stage, we learn the geometric layout changes of the scene while fixing the appearance model and growing sparse Gaussian primitives from COLMAP. Finally, we refine both the appearance model and parameters of added Gaussians.
  } 
  \label{fig:pipeline}
  \vspace{-3mm} 
\end{figure*}



Given a trained 3D Gaussian field, along with time-varying scene images and corresponding poses, our method can continuously update and record changes in this scene. In Sec.~\ref{ssec:formulation}, we briefly introduce rasterization principles and the task formulation of continual learning in a 3D Gaussian field. We developed a three-stage scheme to progressively update the scene representation, avoiding interference from the tight coupling of appearance and geometry properties(Sec.~\ref{ssec:update}). Moreover, in Sec.~\ref{ssec:cl}, we elaborate on our visibility-aware learning technique with generative replay for self-aware scene updating.

\subsection{Preliminaries}
\label{ssec:formulation}
\boldparagraph{Rasterization Principles.} We defines a 3D scene as a set of 3D Gaussian primitives with 3D covariance $\mathbf{\Sigma}\in\mathbb{R}^{3\times3}$, locations $\mathbf{\mu}\in\mathbb{R}^{3}$, opacity $\mathbf{\sigma}$, and spherical harmonics coefficients $\mathbf{Y}$:

\begin{equation}
    \textbf{G} = \{ G_i: (\mu_i, \Sigma_i, \sigma_i, Y_i)~|~ i=1,...,N\}.
    \label{eq:1}
\end{equation}
It is worth noting that, to ensure positive semi-definite in the gradient descent, $\mathbf{\Sigma}$ is formulated using a scaling matrix $\mathbf{S}=diag([s])$ and rotation Matrix $\mathbf{R}$:
\begin{equation}
    \Sigma = RSS^{T}R^{T}.
    \label{eq:4}
\end{equation}
Following EWA Splatting~\cite{zwicker2001ewa}, 
given a world-to-camera rotation $\mathbf{W}$ and the Jacobian $\mathbf{J}$ of the affine approximation of the projective transformation, 3D Gaussians $\{\mathbf{\Sigma}, \mathbf{\mu}\}$ can be projected to 2D distributions $\{\mathbf{\Sigma}', \mathbf{\mu}'\}$ as:
\begin{equation}
    \Sigma' = JW\Sigma W^{T}J^{T}.
    \label{eq:2}
\end{equation}
$\mu'$ indicates the pixel coordinates of Gaussian center points. For view synthesis, the rasterizer sorts all Gaussians in a front-to-back order and performs $\alpha$-blending rendering as follows:
\begin{equation}
    \alpha_i = \sigma_ie^{-\frac{1}{2}(x-\mu')^T\Sigma'^{-1}(x-\mu')},
    \label{eq:3}
\end{equation}
\begin{equation}
    T_i = \prod_{j=1}^{i-1}(1 - \alpha_j),
\end{equation}
\begin{equation}
    C = \sum_{i \in N}c_i\alpha_i T_i,
\end{equation}
where $N$ is the number of 3D Gaussian primitives,  $c_i$  
is the color of Gaussians obtained from spherical harmonics coefficients $\mathbf{Y}$, $\alpha_i$ is the opacity contributed to a pixel, and $T_i$ is the is the accumulated transmittance.\\

\boldparagraph{Task Formulation.} In the context of a time-varying scene and Gaussian representation, we can define the continual learning task in a 3D Gaussian field as follows:
\begin{enumerate}
  \item At newly arrived time step $t$, a set of multi-view images $\mathcal{I}_{gt}=\left\{ I_{f}^{t} \mid f = 1 \dots N \right\}$ of the changed scene, along with camera poses, are collected.
  \item Based on these latest observations, while preserving prior scene knowledge, an updated 3D Gaussian field can be obtained from the one-at-time step $t$-1, enabling continual updating and scene recall at different times.
  \item A flexible and continual 3D Gaussian field is built up to achieve photo-realistic novel view synthesis and support the visualization of changes in a time-varying scene.
\end{enumerate}
Unlike short-term continuous dynamics~\cite{4d-gs, deformable-gs, 3dgstream, d-nerf, zhao2024gaussianprediction}, our goal is to follow temporally arbitrary changes in a scene over a long term, which requires adaptive and self-aware Gaussian management 
in 3D Gaussian fields. As shown in Figure~\ref{fig:pipeline}, a modularized approach has been proposed to effectively handle this task, and we will detail different modules below.

\subsection{Model Update }
\label{ssec:update}

In order to temporally model scene changes, we additionally introduce an efficient hash grid into our 3D Gaussian field as a global appearance model, generating a hybrid scene representation. In the real world, environment changes mainly occur in two aspects: global illumination and scene layout. Straightforward optimization for both of them, especially in a $\alpha$-blending method, will cause serious ambiguity and overfitting problems.
Therefore, we design a progressive solution to decouple lighting and layout parts to achieve accurate and fine-grained scene updates.

\subsubsection{First Stage: Global Appearance Update}
\label{sssec:stage1}
\boldparagraph{Global Appearance Model.}As shown in Figure~\ref{fig:pipeline}, to model the global appearance changes caused by lighting, a 4D hash grid $\mathbf{H}$ and a tiny MLP $\mathbf{F}$ are used to serve as a global appearance model to infer incremental scaling and spherical harmonics properties $\{\Delta s_i^t, \Delta Y_i^t\}$ on Gaussian primitives at different times: 
\begin{equation}
    \{\Delta s_i^t, \Delta Y_i^t\} = F(H\{\mu_i, t\}).
\end{equation}

\boldparagraph{Appearance Updates in Layout-invariant Areas.} In general, global illumination dominates appearance changes in layout-invariant regions (i.e., the static regions without any geometric changes). In the first stage, our approach focuses on updating such changes by separately optimizing the 4D hash grid. In order to avoid the influence caused by the possible new or removing objects in the scene, we design a robust method to determine the layout-invariant areas. Specifically, suppose $I^t_f$ is the newly captured image, we can render $I^{t-1}_f$ with the camera pose of $I^t_f$ and the previous neural model. Then for this pair of corresponding frames $\{I^t_f, I^{t-1}_f\}$, we can acquire their instance segmentation masks $\{S^t_f, S^{t-1}_f\}$ from a foundation model SAM~\cite{sam}. Furthermore, we can calculate the Intersection over Union (IoU) score between corresponding instances from $\{S^t_f, S^{t-1}_f\}$. Layout-invariant areas with high IoU scores at both time steps $t$ and $t$-1 will be masked into $\{M^t_f, M^{t-1}_f\}$. At the current time step $t$, a final layout-invariant mask $M_f$ comes from the intersection of $M^t_f$ and $M^{t-1}_f$, and it will be used to guide subsequent optimization of the appearance model.  Combined with this mask $M_f$ and photometric loss (Eq.~\ref{eq:st-1}-\ref{eq:st-3}), the hash-encoding appearance model can be accurately optimized in layout-invariant areas.
\begin{equation}
    \mathcal{L}_{st} = (1-\lambda)\mathcal{L}_1 \cdot \mathcal{M}_{f} + \lambda \mathcal{L}_{D-SSIM} \cdot \mathcal{M}_{f},
    \label{eq:st-1}
\end{equation}
\begin{equation}
    \mathcal{L}_1 = \frac{1}{HW}\sum_{n=1}^{HW}|C_n - \hat{C}_n|,
    \label{eq:st-2}
\end{equation}
\begin{equation}
    \mathcal{L}_{D-SSIM} = SSIM(C, \hat{C}).
    \label{eq:st-3}
\end{equation}

By excluding all possible dynamic objects or parts, we can learn the global appearance model better, which further benefits the following geometric layout update.

\subsubsection{Second Stage: Geometric Layout Update}
\label{sssec:stage2}

In areas where the geometric layout changes, it is necessary to prioritize growing and pruning Gaussian primitives to fit the correct geometry shape and avoid unexpected overfitting. For emerging objects, new Gaussian primitives $G_{add}$ are spawned from the COLMAP and added to the existing Gaussian field. For missing objects, we set learnable removal factors $m$ which are activated by a function $\psi$ as in Eq.~\ref{eq:nd-2}. Notably, $\psi(m)$ is only multiplied by the opacity of previous Gaussians $G_p$. 

At this stage, the removal factors $m$ and properties of $G_{add}$ are optimized as in Eq.~\ref{eq:nd-1} while keeping the global appearance model learned in the previous stage fixed. The regularization term is defined in Eq.~\ref{eq:nd-3} to drive $\psi(m)$ towards 0 or 1.\\
\begin{equation}
    \mathcal{L}_{nd} = (1-\lambda_1)\mathcal{L}_1+ \lambda_1 \mathcal{L}_{D-SSIM} + \mathcal{L}_{reg} ,
    \label{eq:nd-1}
\end{equation}
\begin{equation}
    \psi(m) = \frac{1}{1 + e^{-1000m}},
    \label{eq:nd-2}
\end{equation}
\begin{equation}
    \mathcal{L}_{reg} = \lambda_2(1-\psi(m))\psi(m)+\lambda_3BCE(\psi(m),1).
    \label{eq:nd-3}
\end{equation}
Ultimately, we set a threshold $\tau=0.01$ to identify candidate Gaussian primitives $G_c$ ($\psi(m)<\tau$) to be removed. Moreover, we further employ the DBSCAN~\cite{dbscan} algorithm to filter out sparse outliers in $G_c$ which are far from surfaces of missing objects, and the remaining Gaussians will be clustered into a bounding box where any Gaussians are pruned from $G_p$. Different from the implicit neural field, our explicit continual learning method allows pruned Gaussians to be stored in a visibility pool (Sec.~\ref{sssec:stage3}) for visualizing changes and recalling previous scene models.

\subsubsection{Third Stage: Joint Refinement}
\label{sssec:stage3}


After pruning Gaussian primitives, rendering artifacts inevitably appear around removed objects. Thus, in the third stage, we perform a joint refinement, following the photometric loss in Eq.~\ref{eq:rd-1},  to optimize the appearance model and properties of $G_{add}$ together.
\begin{equation}
    \mathcal{L}_{rd} = (1-\lambda)\mathcal{L}_1+ \lambda \mathcal{L}_{D-SSIM}.
    \label{eq:rd-1}
\end{equation}
\boldparagraph{Importance Purning.} Inspired by \cite{radsplat}, we expect to extract valuable Gaussians to reduce memory consumption. By traversing all training views at the current time step, we can calculate an importance score $v_i$ of the $i$-th Gaussian as follows: 
\begin{equation}
    v_i = \max_{n \in N_1} (\alpha_i^n T_i^n),
    \label{eq:rd-2}
\end{equation}
where $N_1$ represents the set of pixels related to the $i$-th Gaussian in all our training views. The Gaussians with importance scores less than 0.05 will also be discarded.
\subsection{Visibility-aware Continual Learning}
\label{ssec:cl}

A continual 3D Gaussian field can not only solve the catastrophic forgetting problem over the long term but also recall the scene representation at any previous time step. To this end, we maintain a visibility pool $P_v$ for each Gaussian in the entire time period. Additionally, we store the camera extrinsic of historical training views for generative play~\cite{shin2017continual}.

\boldparagraph{Visibility Pool.}
The explicit scene representation clearly indicates Gaussian primitives on which scene changes occur, which is a significant advantage over existing implicit works, especially in continual scene updates. In our method, we set a visibility pool to store all involved Gaussians over time. Pruned Gaussians will actually be set to inactive, and vice versa. This novel visibility-aware strategy allows our method to perform primitive-level control over the scene map, and it also serves as the foundation for visualizing scene changes. 


\boldparagraph{Generative Replay.}
Benefiting from efficient and photo-realistic view synthesis, Gaussian-based scene representation is naturally suitable for generative replay~\cite{shin2017continual} in continual learning, which only requires recording a small number of camera extrinsic. Specifically, at each new time step, based on previous camera poses and trained 3D Gaussian fields, we can re-render past views as historical cues in our training dataset to mitigate the forgetting problem.


\section{Experiments}
\label{sec:Exp}


{
\renewcommand{\arraystretch}{1.5} 
\begin{table*}[htbp]

\centering
    \scriptsize
    \begin{tabular}{lcccccccccc}
        \Xhline{1.2pt}
      & \multicolumn{2}{c}{\textbf{Breville}}        & \multicolumn{2}{c}{\textbf{Kitchen}} & \multicolumn{2}{c}{\textbf{Living room}}    & \multicolumn{2}{c}{\textbf{Community}} & \multicolumn{2}{c}{\textbf{Spa}} \\ \cmidrule(lr){2-3} \cmidrule(lr){4-5} \cmidrule(lr){6-7} \cmidrule(lr){8-9} \cmidrule(lr){10-11} 
        \multirow{-2}{*}{Method}  & PSNR($\uparrow$)     & SSIM($\uparrow$)     & PSNR($\uparrow$)    & SSIM($\uparrow$)      & PSNR($\uparrow$)   & SSIM($\uparrow$)    & PSNR($\uparrow$)     & SSIM($\uparrow$) 
        & PSNR($\uparrow$)     & SSIM($\uparrow$) 
        \\ \toprule
        Baseline & 20.66 & 0.745  & 16.44 & 0.610
         & 17.28 & 0.720  & 11.13 & 0.448 & 15.38 & 0.580   \\

        CLNeRF~\cite{cl-nerf-iccv}  & 28.02 & 0.826  & \cellcolor[HTML]{F59194}28.40 & 0.877
         & \cellcolor[HTML]{FAC791}24.58 & 0.829  & 22.88 & 0.629 & \cellcolor[HTML]{FAC791}26.28 & 0.811   \\ 
        
         
        4DGS~\cite{4d-gs}     & \cellcolor[HTML]{FAC791}28.92 & \cellcolor[HTML]{FAC791}0.908  & 27.03 & \cellcolor[HTML]{FAC791}0.900 & 24.41 & \cellcolor[HTML]{FAC791}0.873  & \cellcolor[HTML]{FAC791}22.99 & \cellcolor[HTML]{FAC791}0.711 & 26.04 & \cellcolor[HTML]{FAC791}0.859  \\
        

        Ours    & \cellcolor[HTML]{F59194}30.11 & \cellcolor[HTML]{F59194}0.927  & \cellcolor[HTML]{FAC791}28.02 & \cellcolor[HTML]{F59194}0.912
         & \cellcolor[HTML]{F59194}26.10 & \cellcolor[HTML]{F59194}0.881  & \cellcolor[HTML]{F59194}23.88 & \cellcolor[HTML]{F59194}0.764 & \cellcolor[HTML]{F59194}27.84 & \cellcolor[HTML]{F59194}0.891  \\

         \hline

         UB   & 30.36 & 0.929  & 27.99 & 0.911
         & 26.22 & 0.882  & 23.88 & 0.755 & 28.16 & 0.894  \\
         
        \Xhline{1.2pt}

      & \multicolumn{2}{c}{\textbf{Street}}        & \multicolumn{2}{c}{\textbf{Car}} & \multicolumn{2}{c}{\textbf{Grill}}    & \multicolumn{2}{c}{\textbf{Mac}} & \multicolumn{2}{c}{\textbf{Ninja}}  \\ \cmidrule(lr){2-3} \cmidrule(lr){4-5} \cmidrule(lr){6-7} \cmidrule(lr){8-9} \cmidrule(lr){10-11} 
        \multirow{-2}{*}{Method}  & PSNR($\uparrow$)     & SSIM($\uparrow$)     & PSNR($\uparrow$)    & SSIM($\uparrow$)      & PSNR($\uparrow$)   & SSIM($\uparrow$)    & PSNR($\uparrow$)     & SSIM($\uparrow$) 
        & PSNR($\uparrow$)     & SSIM($\uparrow$) 
        \\ \toprule
        Baseline  & 13.12 & 0.486  & 19.53 & 0.610
         & 19.01 & 0.575  & 17.80 & 0.778 & 18.97 & 0.777   \\ 
        
         CLNeRF~\cite{cl-nerf-iccv}  & \cellcolor[HTML]{FAC791}22.53 & 0.612  & 22.73 & 0.541
         & 24.48 & 0.653  & 29.33 & 0.906 & 27.19 & 0.878   \\ 
        

        4DGS~\cite{4d-gs}      & 22.26 & \cellcolor[HTML]{FAC791}0.680  & \cellcolor[HTML]{FAC791}23.13 & \cellcolor[HTML]{FAC791}0.662 & \cellcolor[HTML]{FAC791}24.94 & \cellcolor[HTML]{FAC791}0.733  & \cellcolor[HTML]{FAC791}29.65 & \cellcolor[HTML]{FAC791}0.930 & \cellcolor[HTML]{FAC791}27.64 & \cellcolor[HTML]{FAC791}0.912  \\
        
        Ours    & \cellcolor[HTML]{F59194}22.80 & \cellcolor[HTML]{F59194}0.703  & \cellcolor[HTML]{F59194}23.81 & \cellcolor[HTML]{F59194}0.742
         & \cellcolor[HTML]{F59194}25.31 & \cellcolor[HTML]{F59194}0.770  & \cellcolor[HTML]{F59194}29.92 & \cellcolor[HTML]{F59194}0.934 & \cellcolor[HTML]{F59194}27.75 & \cellcolor[HTML]{F59194}0.921  \\


         \hline

         UB   & 23.23 & 0.715  & 23.77 & 0.741
         & 25.46 & 0.774  & 30.17 & 0.935 & 27.98 & 0.921  \\
        \Xhline{1.2pt}
    \end{tabular}
    
\vspace{-3mm} 
\caption{\textbf{Quantitative comparison on WAT dataset}. Best results are highlighted as \fs{\bf{first}},~\nd{\bf{second}}. }
    \label{tab:WAT_quantitative}
    \vspace{-5mm} 
\end{table*}
}

\subsection{Dataset and Implementation details}
\boldparagraph{Dataset.}
We evaluate \method on the World Across Time (WAT) dataset~\cite{cl-nerf-iccv}, which comprises 10 real scenes encompassing both indoor and outdoor environments. Additionally, we conducted experiments on the Synthetic NeRF dataset~\cite{nerf}, which comprises 8 distinct scenes.

We adopted the Synthetic NeRF dataset partitioning strategy from CLNeRF~\cite{cl-nerf-iccv}, dividing the training images of each scene into 10 sequential, equally-sized subsets. During training, these subsets were used sequentially as ground truth images.
Although the scenes are static, the varying viewpoints change across subsets.
To prevent forgetting, we employed a generative replay approach to generate historical images for supervision, utilizing saved camera poses. Since each scene in the Synthetic NeRF dataset theoretically remains static, we followed CLNeRF~\cite{cl-nerf-iccv} by omitting appearance embeddings, as well as the appearance model and visibility pool for this dataset. 

\boldparagraph{Implementation details.} Our approach updates a given 3D Gaussian field using a three-stage training strategy. Initially, we initialize the 3D Gaussian field by leveraging the dense points from MVS~\cite{mvs} to achieve a more refined Gaussian field initialization. In the first stage of updating the Gaussian field, we conduct 7k iterations of training, focusing solely on the global appearance model, while the densify operation~\cite{3D-Gaussian} is closed.
After the first phase of training, we add the sparse points obtained by SFM~\cite{sfm} as the initialization Gaussians of the new object and open the densify operation. A total of 8k iterations are trained in the second stage. We will use DBSCAN to filter the points to be deleted when 5k iterations and 8k iterations are trained in the second stage.
In the third stage, we conduct a total of 15k training iterations and employ importance pruning to eliminate excess Gaussians in $G_{e}$ after completing 4k iterations of training during this stage. Furthermore, we normalize the time steps $t$ to the $[0, 1]$ for each scene.
All experiments are conducted and evaluated on an NVIDIA GeForce RTX 3090 GPU with 24 GB of RAM.
\begin{figure}[!t]
    \centering
    \begin{subfigure}{0.49\linewidth}
        \centering
        \includegraphics[width=0.99\linewidth]{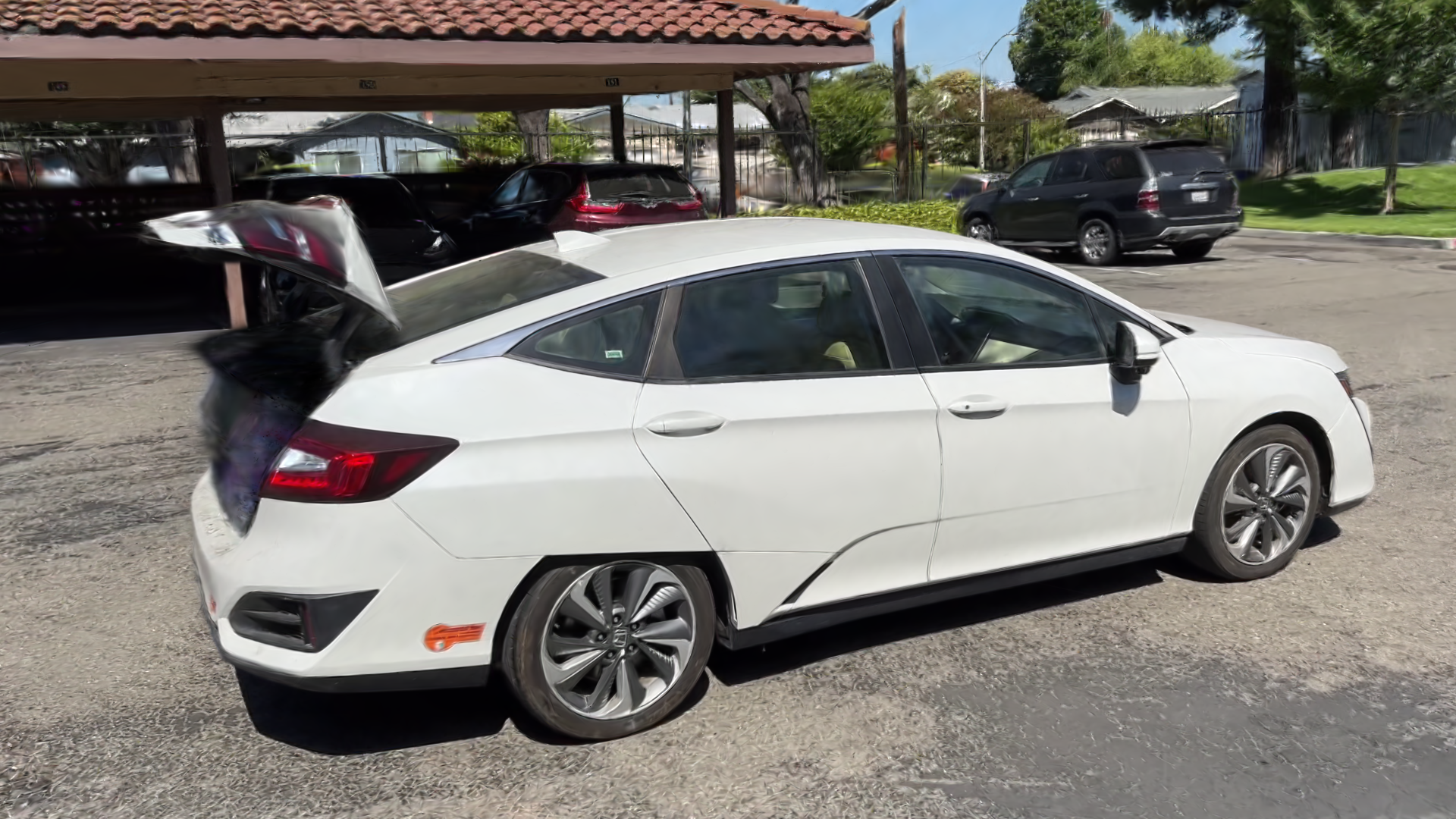}
        \caption{w/o Layout-invariant Mask.}
        \label{subfig:ablation_woMask}
    \end{subfigure}
    \begin{subfigure}{0.49\linewidth}
        \centering
        \includegraphics[width=0.99\linewidth]{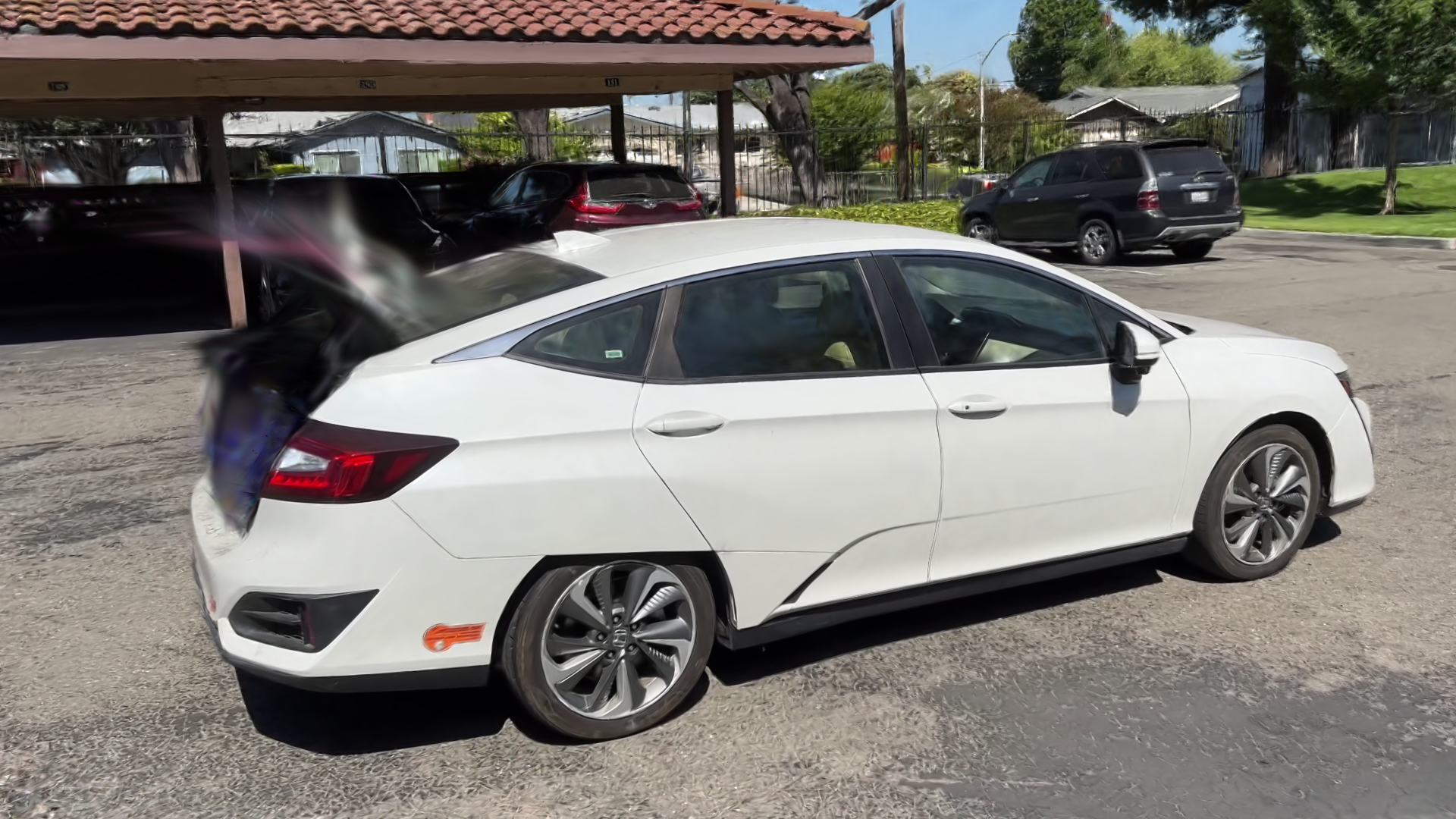}
        \caption{w/o Addition .}
        \label{subfig:ablation_woAdaptive}
    \end{subfigure}
    \begin{subfigure}{0.49\linewidth}
        \centering
        \includegraphics[width=0.99\linewidth]{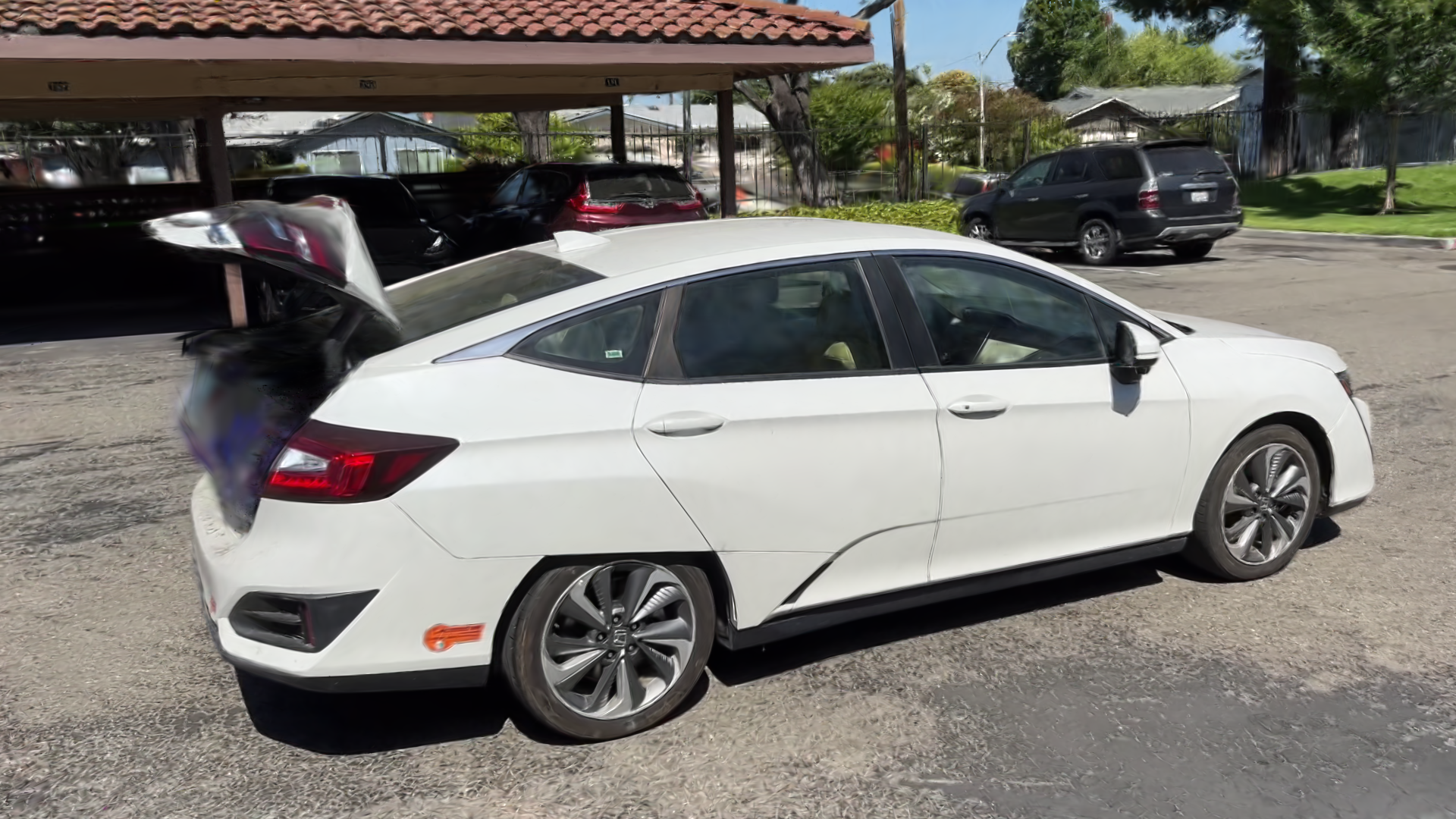}
        \caption{w/o Removal Factors.}
        \label{subfig:ablation_woClusterKnn}
    \end{subfigure}
    \begin{subfigure}{0.49\linewidth}
        \centering
        \includegraphics[width=0.99\linewidth]{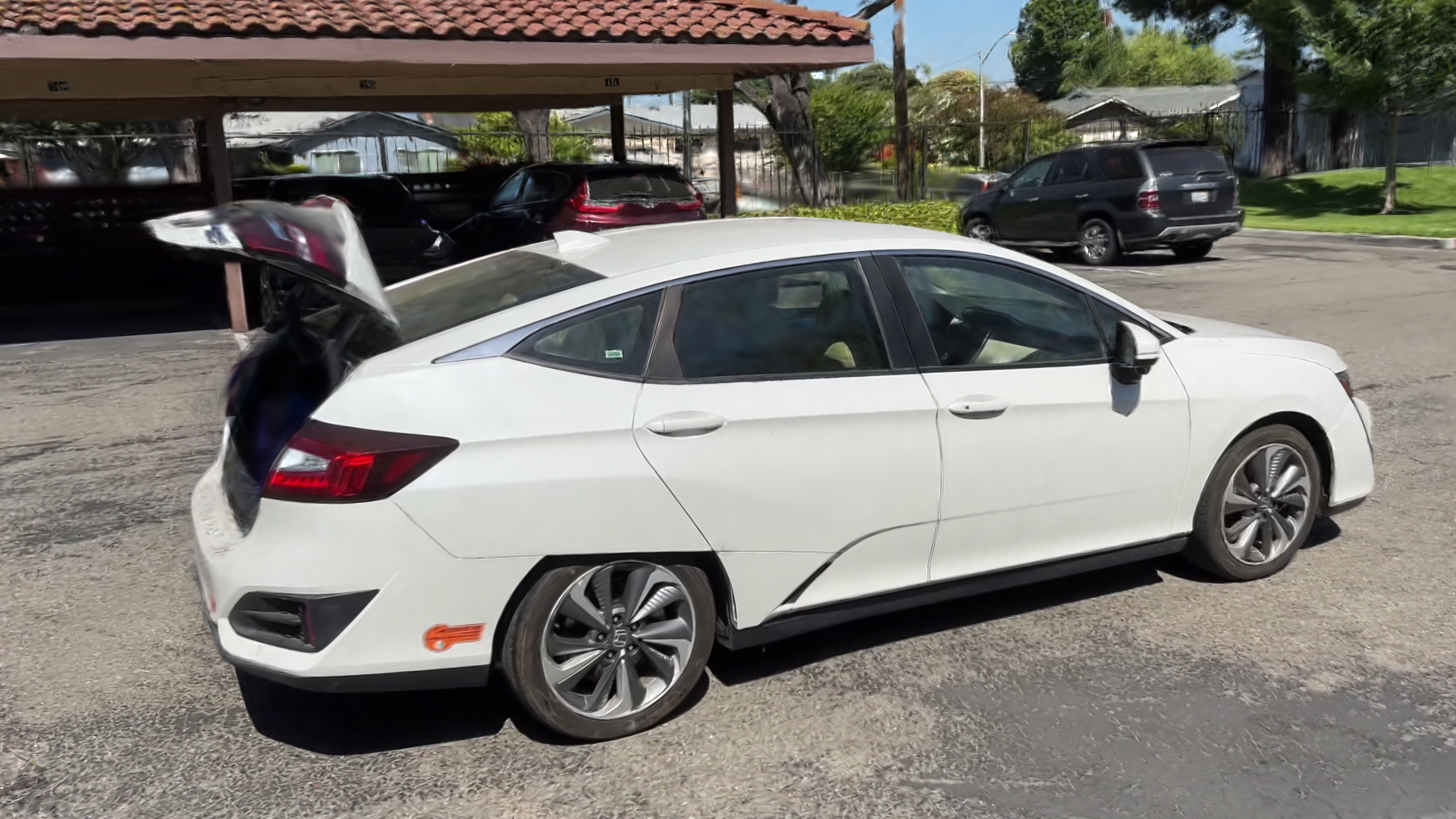}
        \caption{Full Model.}
        \label{subfig:ablation_full}
    \end{subfigure}
    \vspace{-3mm} 
    \caption{\textbf{Effectiveness analysis of each component}.}
    \label{fig:ablation}
    \vspace{-5mm} 
\end{figure}


{
\renewcommand{\arraystretch}{1.5} 
\begin{table*}[htbp]
\centering
    \scriptsize
    \setlength{\tabcolsep}{2pt}  
    \begin{tabular}{lcccccccccccccccc}
        \Xhline{1.2pt}
      & \multicolumn{2}{c}{\textbf{Lego}}        & \multicolumn{2}{c}{\textbf{Chair}} & \multicolumn{2}{c}{\textbf{Drums}}    & \multicolumn{2}{c}{\textbf{Ficus}} & \multicolumn{2}{c}{\textbf{Hotdog}} & \multicolumn{2}{c}{\textbf{Materials}} & \multicolumn{2}{c}{\textbf{Mic}} & \multicolumn{2}{c} {\textbf{Ship}}\\ 
      \cmidrule(lr){2-3} \cmidrule(lr){4-5} \cmidrule(lr){6-7} \cmidrule(lr){8-9} \cmidrule(lr){10-11} \cmidrule(lr){12-13} \cmidrule(lr){14-15} \cmidrule(lr){16-17}   
        \multirow{-2}{*}{Method}  & PSNR($\uparrow$)     & SSIM($\uparrow$)     & PSNR($\uparrow$)    & SSIM($\uparrow$)      & PSNR($\uparrow$)   & SSIM($\uparrow$)    & PSNR($\uparrow$)     & SSIM($\uparrow$) 
        & PSNR($\uparrow$)     & SSIM($\uparrow$) 
        & PSNR($\uparrow$)     & SSIM($\uparrow$) 
        & PSNR($\uparrow$)     & SSIM($\uparrow$) 
        & PSNR($\uparrow$)     & SSIM($\uparrow$) 
        \\ \toprule
        Baseline   & 17.98 & 0.738  & 23.35 & 0.889
         & 18.75 & 0.784  & 26.44 & 0.928 & 28.54 & 0.926
         & 17.44 & 0.738  & 27.17 & 0.935 & 18.56 & 0.583\\

        CLNeRF~\cite{cl-nerf-iccv}  & \cellcolor[HTML]{FAC791}34.80 & \cellcolor[HTML]{FAC791}0.976  & \cellcolor[HTML]{FAC791}34.39 & \cellcolor[HTML]{FAC791}0.980
         & 25.46 & 0.931  & 33.10 & 0.980 & \cellcolor[HTML]{FAC791}36.42 & \cellcolor[HTML]{FAC791}0.979
         & 29.12 & 0.943  & \cellcolor[HTML]{FAC791}34.68 & \cellcolor[HTML]{FAC791}0.987 & 29.30 & 0.880\\ 
        

        4DGS~\cite{4d-gs}      & 33.23 & 0.968  & 34.35 & 0.976
         & \cellcolor[HTML]{FAC791}25.86 & \cellcolor[HTML]{FAC791}0.947  & \cellcolor[HTML]{FAC791}34.77 & \cellcolor[HTML]{FAC791}0.985 & 36.16 & 0.977 
         & \cellcolor[HTML]{FAC791}29.16 & \cellcolor[HTML]{FAC791}0.952  & 34.44 & 0.988 & \cellcolor[HTML]{FAC791}29.48 & \cellcolor[HTML]{FAC791}0.886  \\


        Ours    & \cellcolor[HTML]{F59194}35.70 & \cellcolor[HTML]{F59194}0.980  & \cellcolor[HTML]{F59194}35.21 & \cellcolor[HTML]{F59194}0.986
         & \cellcolor[HTML]{F59194}26.07 & \cellcolor[HTML]{F59194}0.951  & \cellcolor[HTML]{F59194}34.98 & \cellcolor[HTML]{F59194}0.985 & \cellcolor[HTML]{F59194}37.59 & \cellcolor[HTML]{F59194}0.983
         & \cellcolor[HTML]{F59194}30.205 & \cellcolor[HTML]{F59194}0.958  & \cellcolor[HTML]{F59194}36.33 & \cellcolor[HTML]{F59194}0.991 & \cellcolor[HTML]{F59194}31.47 & \cellcolor[HTML]{F59194}0.903\\

         \hline

         UB   & 36.27 & 0.982  & 35.74 & 0.987
         & 26.32 & 0.954  & 35.52 & 0.987 & 38.12 & 0.985  
         & 30.52 & 0.960  & 36.68 & 0.992 & 31.79 & 0.906\\
         
        \Xhline{1.2pt}

    \end{tabular}
    \vspace{-3mm}
\caption{\textbf{Quantitative comparison on Synthetic NeRF dataset}. Best results are highlighted as \fs{\bf{first}},~\nd{\bf{second}}.}

    \label{tab:nerfsynthetic_quantitative}
    \vspace{-3mm}
\end{table*}
}
\begin{figure*}[htbp]
  \centering
  \includegraphics[width=0.98\linewidth, trim={0 0 0 0}, clip]{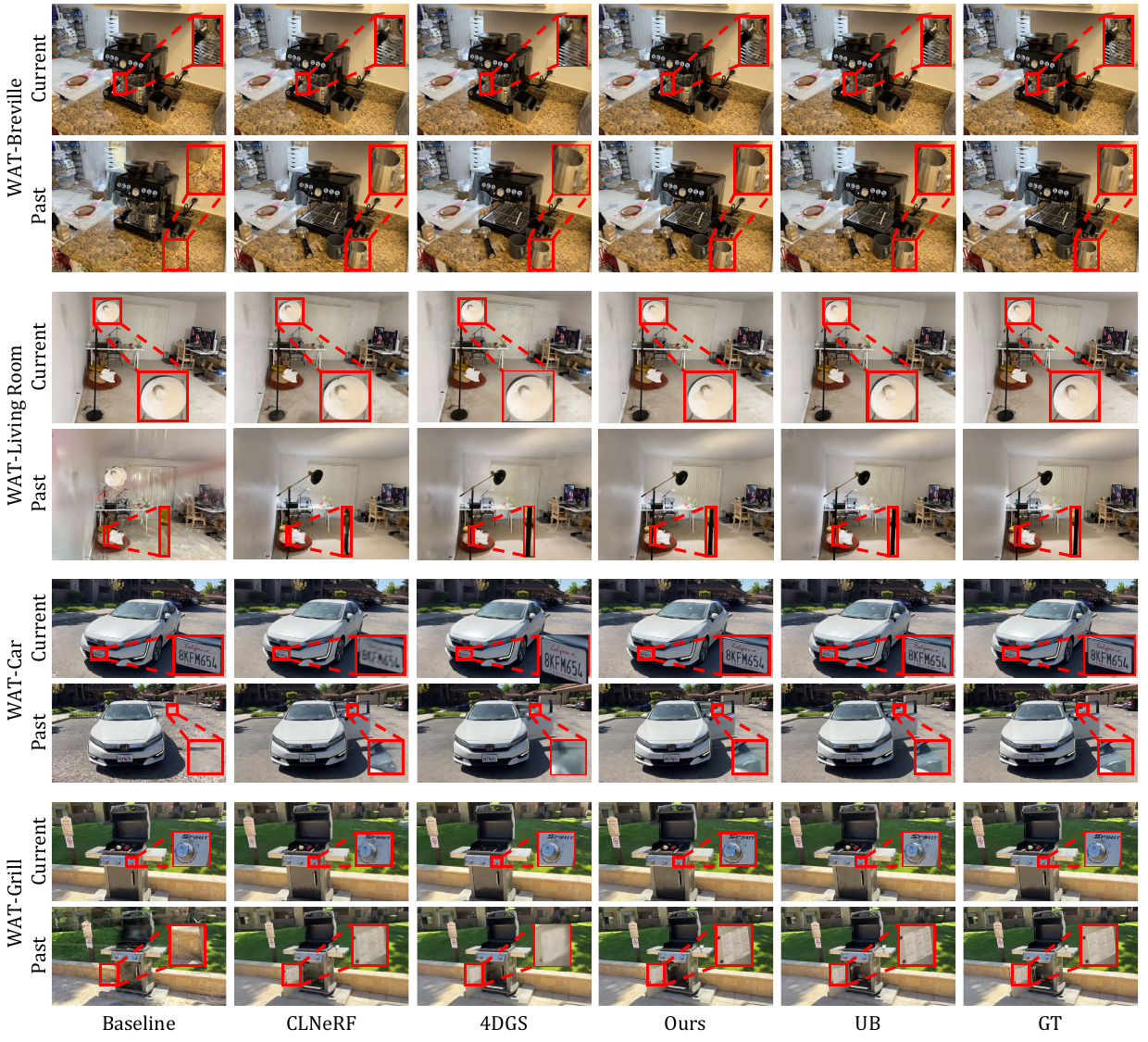}
  \vspace{-3mm}
  \caption{\textbf{Visualizaton results on WAT dataset}. Each pair of lines shows the current and past test images rendered by different methods. NT overfitted the current data, leading to forgetting past knowledge. Our method demonstrates better rendering results compared to CLNeRF and 4DGS.}
  \vspace{-3mm}
  \label{fig:visualization_results}
\end{figure*}


\begin{figure}[ht]
    \centering
    \begin{subfigure}{0.49\linewidth}
        \centering
        \includegraphics[width=0.99\linewidth]{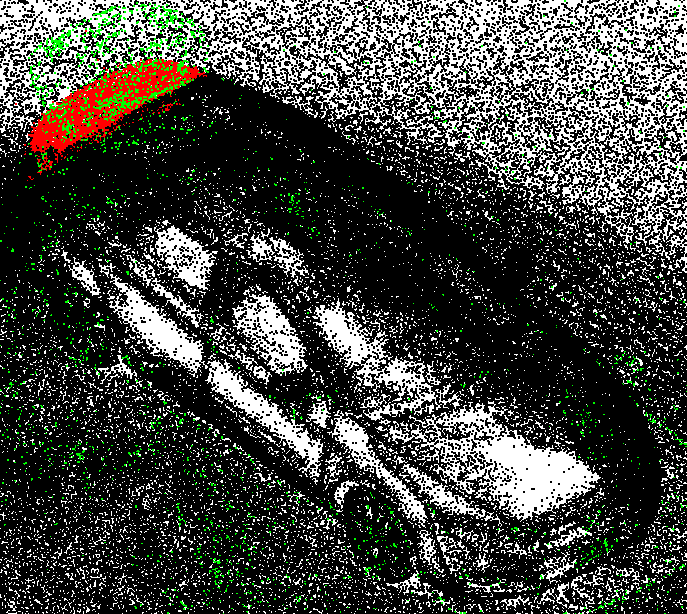}
        \caption{$\mathbf{T}_1$}
        \label{subfig:ablation_pcd_woMask}
    \end{subfigure}
    \begin{subfigure}{0.49\linewidth}
        \centering
        \includegraphics[width=0.99\linewidth]{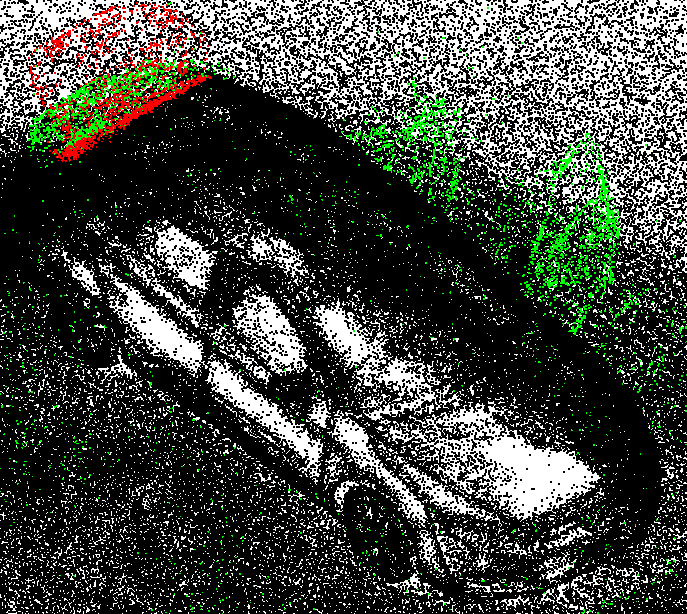}
        \caption{$\mathbf{T}_2$}
        \label{subfig:ablation_pcd_woAdaptive}
    \end{subfigure}
    \begin{subfigure}{0.49\linewidth}
        \centering
        \includegraphics[width=0.99\linewidth]{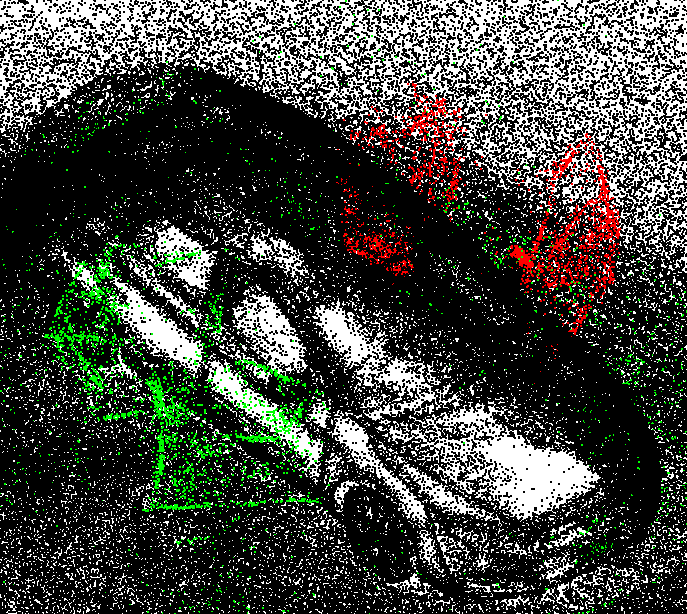}
        \caption{$\mathbf{T}_3$}
        \label{subfig:ablation_pcd_woClusterKnn}
    \end{subfigure}
    \begin{subfigure}{0.49\linewidth}
        \centering
        \includegraphics[width=0.99\linewidth]{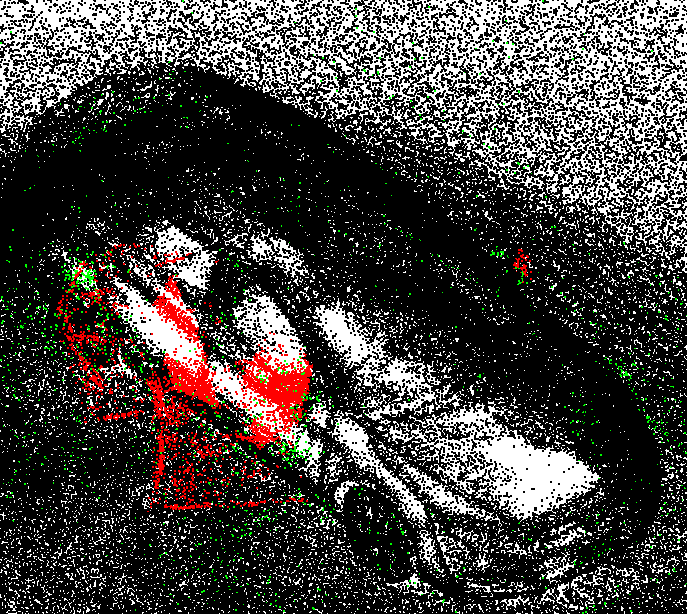}
        \caption{$\mathbf{T}_4$}
        \label{subfig:ablation_pcd_full}
    \end{subfigure}
    \vspace{-2mm} 
    \caption{\textbf{The visualization results of the scene geometric changes learned by ours in scene car at each time}. The green points represent the objects that have been added compared to the previous time, while the red points indicate the objects that disappeared compared to the previous time.}
    \vspace{-2mm} 
    \label{fig:ablation_pcd}
\end{figure}

\subsection{Evaluation}

To comprehensive evaluation, we compare \method to various methods for changing environments.
(1) Baseline: the baseline model refers to 3D Gaussian Splatting~\cite{3D-Gaussian} combined with our proposed appearance model and visibility pool, without using generative replay data. The baseline represents the lower bound performance of our model on continuous updates.
(2) CLNeRF~\cite{cl-nerf-iccv}: the SOTA method for continual learning with NeRF; \footnote{CL-NeRF \cite{cl-nerf-nips} is also a NeRF-based continual learning method, while the code is not publicly available at the time of our submission.} 
(3) 4DGS~\cite{4d-gs}: 4DGS extends 3DGS to handle dynamic scenes by incorporating temporal dynamics into the differentiable Gaussian rendering process. This approach predicts a 4D joint distribution that captures the time-variant behavior of Gaussian primitives;
Additionally, 4DGS necessitates inputting data from all time points simultaneously to learn the 4D Gaussian distribution effectively.
(4) Upper bound model (UB): we update the 3D Gaussian field by replacing generative replay with memory reply~\cite{chaudhry2019tiny, lopez2017gradient} (data memory pool stores all past GT images) to represent the upper bound our method can achieve.


As shown in Table~\ref{tab:WAT_quantitative} and Table~\ref{tab:nerfsynthetic_quantitative}, our method generally outperforms CLNeRF across most scenes, except for the kitchen scene in the WAT dataset where PSNR is slightly lower. Notably, our method consistently achieves significantly higher SSIM scores across all scenes, suggesting that our rendered images more closely resemble the ground truth images in the test set.  Moreover, leveraging the inference speed of 3DGS, our method achieves real-time rendering speeds of 34 FPS on the WAT dataset, while CLNeRF lags significantly behind at 0.75 FPS. 
Furthermore, our method consistently outperforms 4DGS across all scenes. Notably, since the Synthetic NeRF dataset represents a static scene, the time input for all images is set to 1 when training 4DGS on this dataset.
In comparison to UB, which utilizes memory replay, our method achieves comparable rendering quality with lower storage consumption, as it does not necessitate storing historical images.

\begin{table}[t!]
    \centering
    \setlength{\extrarowheight}{2pt}
    \scalebox{0.9}{
    \begin{tabular}{lcc}
    \toprule
    \multicolumn{1}{c}{\multirow{2}{*}{Config.}} & \multicolumn{2}{c}{WAT} \\ \cmidrule(lr){2-3} 
    \multicolumn{1}{c}{} & \multicolumn{1}{l}{PSNR $\uparrow$} & \multicolumn{1}{l}{SSIM $\uparrow$}   \\ \hline
    w/o Removal Factors & 25.87  & 0.834\\
    w/o Addition of 3D Gaussians & 25.96 & 0.829 \\

    w/o Layout-Invariant Mask & \cellcolor[HTML]{FAC791}26.18 & \cellcolor[HTML]{FAC791}0.839  \\
    Full Model & \cellcolor[HTML]{F59194}26.55  & \cellcolor[HTML]{F59194}0.844  \\
    \bottomrule
    \end{tabular}
    }
    \vspace{-2mm} 
    \caption{
    \textbf{Ablation study of each step discussed in Sec.~\ref{ssec:update} on WAT dataset.} Best results are highlighted as\fs{\bf{first}},~\nd{\bf{second}}.
    }
    \label{tab:ablation}
    \vspace{-2mm}
\end{table}



The visualization results in Figure~\ref{fig:visualization_results} demonstrate that our method outperforms CLNeRF in handling details. Baseline overfits the current data, resulting in catastrophic forgetting of previous images.
Since each scene in the WAT dataset undergoes significant changes over time, 4DGS is unable to accurately reconstruct the scene at each moment, even when data from all time points is provided during training. 
Table~\ref{table:first_five_moments} illustrates the average PSNR of our method across sequential updates from Time1 to Time4 on the WAT dataset (detailed data for each scene is available in the supplementary file). As scenes are updated, the knowledge acquired from past scenes is largely retained. Compared to the baseline, our method effectively mitigates catastrophic forgetting.

Our approach benefits from the advantages offered by 3DGS~\cite{3D-Gaussian} for explicit representation, enabling it to detect alterations in scene geometric layout. As depicted in Figure~\ref{fig:ablation_pcd}, our method effectively discerns variations in scene geometry resulting from the act of opening and closing the car door. While CLNeRF uses an implicit radiance field to represent the scene, it lacks the ability to capture geometric changes in the scene. Please refer to our supplementary material for more experimental results.

\subsection{Ablation Study}
\paragraph{Removal Factors.}
We begin by analyzing the consequences of not learning the removal factors on our results. As illustrated in Figure \ref{fig:ablation}, 
without learning removal factors result in the persistence of objects that should have disappeared, thereby producing artifacts. Additionally, the quantification in Table~\ref{tab:ablation} demonstrates the effectiveness of learning the removal factors in deleting disappearing objects.

\begin{table}[t!]
    \centering
    \setlength{\extrarowheight}{2.2pt}
    \scalebox{0.9}{
    \begin{tabular}{c|c c c c c}
        \toprule
        \multicolumn{2}{c}{\multirow{2}{*}{Testing on}} & \multicolumn{4}{c}{WAT} \\
        \cmidrule(lr){3-6}
        \multicolumn{2}{c}{} & $\mathbf{T}_1$ & $\mathbf{T}_2$ & $\mathbf{T}_3$ & $\mathbf{T}_4$ \\ \hline
        \multirow{4}{*}{\rotatebox{90}{Training on}}
        & $\mathbf{T}_1$ & 26.92 & - & - & - \\
        & $\mathbf{T}_2$ & 26.90 & 26.05 & - & - \\
        & $\mathbf{T}_3$ & 26.86 & 25.99 & 26.68 & - \\
        & $\mathbf{T}_4$ & 26.84 & 25.92 & 26.61 & 26.42 \\
        \toprule
    \end{tabular}
    }
    \vspace{-2mm} 
    \caption{\textbf{Average rendering results from $\mathbf{T}_1$ to $\mathbf{T}_4$ for sequential updates on WAT Dataset}.}
    \label{table:first_five_moments}
\end{table}

\boldparagraph{Self-aware Addition of 3D Gaussians.}
In the second stage of model updating, we learn the newly added objects by incorporating sparse point initialization and densifying only the newly added points. To analyze the effectiveness of adding sparse points, we conducted experiments where we solely relied on the densify operation to learn new objects. As shown in Figure~\ref{fig:ablation} and Table~\ref{tab:ablation}, using only the densified operation of 3DGS to add points does not sufficiently learn the newly appearing objects. However, leveraging the rough geometric cues from the sparse point cloud leads to improved results.


\boldparagraph{Layout-invariant Mask.}
We also analyzed whether to use the layout-invariant mask. As shown in Figure~\ref{fig:ablation} and Table~\ref{tab:ablation}, neglecting the layout-invariant mask can cause the model to 
adapt to the visual alterations in the geometric change area during the initial training stage, potentially leading to the failure of removal for disappeared objects.

\section{Conclusion}
\label{sec:Conclusion}

This paper presents \method, a novel approach integrating 3D Gaussian Splatting with continual learning. Our method updates the neural model with current data while preserving past information, explicitly modeling various changes through a multi-stage strategy. Additionally, our visibility-aware continual learning with generative replay supports self-aware updates without storing images. Experimental results demonstrate superior rendering quality and effective visualization of changes over time.

\noindent \textbf{Limitation.}
Similar to most existing 3DGS-based methods, \method also fails to accurately model scenes with strong reflections (e.g., mirrors), and updates in these areas result in incorrect geometry or appearance, which may be solved by integrating more advanced 3D representations in the future.


\noindent \textbf{Acknowledgments. }
This work was supported by the National Key R\&D Program of China (Grant No. 2024YFB4505500 \& 2024YFB4505501).
We also express our gratitude to all the anonymous reviewers for their professional and insightful comments.

{
    \small
    \bibliographystyle{ieeenat_fullname}
    \bibliography{main}
}

\end{document}